\documentclass[11pt,a4paper]{article}
\usepackage[margin=25mm]{geometry}
\usepackage{iftex}
\ifPDFTeX
  \usepackage[T1]{fontenc}
  \usepackage[utf8]{inputenc}
  \usepackage{lmodern}
  \input{glyphtounicode}
\else
  \usepackage{fontspec}
\fi
\usepackage{amsmath,amssymb}
\usepackage{graphicx,booktabs,tabularx,array}
\usepackage[font=small,labelfont=bf]{caption}
\usepackage{microtype}
\usepackage{needspace}
\usepackage{titlesec}
\usepackage{xcolor}
\definecolor{linkblue}{RGB}{25,65,100}
\usepackage[colorlinks=true,linkcolor=linkblue,urlcolor=linkblue,unicode]{hyperref}
\hypersetup{pdftitle={Optimizing the Score, Losing Sight of the Task: Reward Hacking Across Weights, Selection, and Prompts},pdfauthor={Vansh Wahi},pdfsubject={Reward hacking across optimization substrates}}
\titleformat{\section}{\large\bfseries}{}{0pt}{}
\titleformat{\subsection}{\normalsize\bfseries}{}{0pt}{}
\titlespacing*{\section}{0pt}{17pt plus 3pt minus 2pt}{8pt}
\titlespacing*{\subsection}{0pt}{12pt plus 2pt minus 1pt}{5pt}

\begin{document}
\begin{center}
{\LARGE\bfseries Optimizing the Score,\\[4pt] Losing Sight of the Task\par}
\vspace{9pt}
{\large Reward Hacking Across Weights, Selection, and Prompts\par}
\vspace{15pt}
{\large Vansh Wahi\par}
\vspace{3pt}
{\small AI Research Engineer\\\href{mailto:v3wahi@uwaterloo.ca}{v3wahi@uwaterloo.ca}\par}
\vspace{7pt}
{\small September 2026\par}
\end{center}
\vspace{5pt}
\begin{abstract}
A higher evaluation score does not always mean a better language model system. When optimization exploits an evaluator's mistakes, measured progress can conceal unchanged or deteriorating task performance. This failure can arise through parameter updates, selection among generated outputs, or revisions to persistent prompts. We develop a comparative framework for reward hacking across these three optimization substrates: weights, selection, and text. Building on the Proxy Compression Hypothesis and research on inference-time and in-context reward hacking, we examine how reachable behavior, optimization budgets, and persistent adaptation shape exposure to proxy error. We formalize a distance-dependent upper bound on evaluator disagreement and a capacity ordering for nested policy classes, then show why distance alone cannot establish a universal ranking of vulnerability. An exact finite-output illustration demonstrates how the location of a scoring defect changes the behavior favored by each method. We also map representative defenses across substrates, identifying which mechanisms transfer directly and which offer only functional analogies. Persistent prompts receive particular attention: their contents are inspectable, but the behavior induced by a small textual change may be difficult to anticipate. The formal analysis, numerical illustration, and published evidence together provide a basis for comparing optimization methods and identifying the conditions under which their defenses transfer. The resulting framework connects optimization choices to verification requirements: reliable improvement depends on controlling accessible failure modes and preserving evidence of task quality independent of the score being optimized.
\end{abstract}
\section{Introduction}\label{introduction}

A language model system receives a higher score after an update. Has it become better at the task, or better at satisfying the evaluator? That distinction is central to any system that uses measured performance to guide its own improvement. A score makes optimization possible, but it also creates an opportunity: behavior that exploits a weakness in the measurement can be rewarded alongside behavior that solves the task.

This opportunity appears in several familiar forms. Parameter training can increasingly favor outputs that a reward model scores highly even as an independent measure of quality declines (Gao et al., 2023). Best-of-n selection can expose a scorer's blind spots by searching a larger pool of candidates (Khalaf et al., 2025). Persistent prompt optimization can encode a scoring shortcut into instructions that are then reused across future inputs. In one production example, a prompt mutation raised a judge's rationale-alignment pass rate from 23.1\% to 80.0\% by adopting its preferred vocabulary, while defect-identification precision remained unchanged (\href{https://arxiv.org/abs/2609.02246}{Wahi, 2026}). The apparent improvement came from matching the evaluator's preferences more closely.

We study these settings together because they share an optimization problem. Each method changes which outputs the system produces or accepts in response to an imperfect reward signal. What differs is how that change is implemented. Weight optimization modifies parameters; selection changes the distribution of returned candidates; prompt optimization revises the instructions, examples, or memory on which the model is conditioned. We call these three implementations \emph{optimization substrates}.

The shared mechanism has precedents in the literature. The Proxy Compression Hypothesis organizes reward hacking around objective compression, optimization amplification, and evaluator--policy co-adaptation, including connections between training-time and inference-time methods (Wang et al., 2026). Research on in-context feedback and iterative self-refinement demonstrates that reward exploitation can also occur without updating model weights (Pan et al., 2024a, 2024b). Our contribution is an explicit comparison of the three substrates, centered on how their reachable behaviors and available controls differ. Persistent text artifacts are especially useful for this comparison because the intervention can be read directly, even when its behavioral effects remain uncertain.

A common explanation is that more expressive optimization permits more severe reward hacking. This intuition captures something important, but it needs a precise interpretation. A method that reaches farther from its starting distribution may encounter larger evaluator errors. It may also move in directions where the evaluator remains accurate. Conversely, a small change can concentrate probability on an exploit already near the baseline. The severity of hacking therefore depends on the relationship between the reachable policies and the evaluator's errors, as well as the resources available to search those policies.

We develop this distinction through three contributions.

\textbf{A common formal setup.} We describe weights, selection, and persistent prompts through budgeted policy classes, proxy and task rewards, and behavioral divergence. The setup separates what a method can represent from what a particular optimizer discovers, and distinguishes evaluator disagreement from an actual decline in task performance.

\textbf{An account of reach and capacity.} We establish a conditional upper bound on the available proxy gap and a separate ordering result for nested policy classes. A counterexample shows why maximum distance alone does not order vulnerability. An exact finite-output illustration then demonstrates how different scoring defects interact with flexible distribution optimization, bounded conditioning, and best-of-n selection.

\textbf{A defense correspondence.} We compare methods for improving evaluation, constraining optimization, preserving independent evidence, and controlling acceptance. Some mechanisms transfer readily across substrates, while others require different assumptions or forms of access. For example, both KL penalties and prompt-edit budgets can limit optimization, but only the former directly controls the behavioral divergence used in our bounds. The comparison makes these differences explicit while retaining the practical value of the analogy.

The framework also helps interpret differing results on prompt optimization and reinforcement learning. GEPA reports strong performance with fewer rollouts in its evaluated settings (Agrawal et al., 2025), while other studies find task-dependent advantages for simpler prompt search or parameter adaptation (Chang and Chen, 2026; Liu et al., 2026). We treat sample efficiency, feedback quality, and reachable behavior as distinct explanations to be tested. Performance in one setting does not establish a universal ranking of the substrates.

This is a framework paper supported by published observations, mathematical analysis, and a reproducible numerical illustration. Its contribution is the comparative account of reachable behavior, proxy error, and defenses across the three substrates. The numerical illustration is computed over explicitly specified distributions; no new language-model experiments are reported. The practical aim is to connect the choice of optimization method to the evidence needed to trust its improvements.

\section{Background and Scope}\label{background-and-scope}

\subsection{The Proxy Compression Hypothesis}\label{the-proxy-compression-hypothesis}

Wang et al.~(2026) propose the Proxy Compression Hypothesis (PCH) as a framework for understanding reward hacking. Its three organizing dynamics are objective compression, optimization amplification, and evaluator--policy co-adaptation.

\textbf{Compression} occurs when an evaluation procedure represents a rich objective through a restricted signal. Human preferences depend on context and trade-offs; a learned reward model summarizes them in a score. A program specification contains many requirements; a test suite checks a subset. An LLM judge receives a rubric, but its verdict also depends on how it interprets and applies that rubric.

Compression is not, by itself, proof that hacking must occur. A scalar score can perfectly represent the ordering needed for a particular task. The concern is the combination of missing or misvalued information and an optimizer capable of exploiting it. For this paper, \emph{proxy error} includes both omitted requirements and incorrect assessments of requirements the evaluator attempts to measure.

\textbf{Amplification} occurs when optimization increasingly favors the proxy's mistakes. An error that is rare under the base model may become common under a selected or adapted policy. The optimizer need not explicitly represent the evaluator's weakness: preferentially retaining outputs with high scores can be enough to concentrate probability on it.

\textbf{Co-adaptation} concerns repeated changes to both policy and evaluator. A policy exploits the current evaluator; the evaluator or oversight procedure is repaired; subsequent optimization responds to the repair. This differs from adaptation to a fixed evaluator, where the policy changes but the evaluator does not. Both are relevant, and we keep them separate in \hyperref[shared-mechanisms-and-different-dynamics]{Shared Mechanisms and Different Dynamics}.

PCH already discusses inference-time selection. We build on that connection by treating weights, selection procedures, and persistent text artifacts as alternative ways of constructing the evaluated policy. The emphasis is on differences in feasible behavior, inspection, and control, rather than a claim that the underlying mechanism was previously unknown.

\subsection{What the three-substrate abstraction covers}\label{what-the-three-substrate-abstraction-covers}

The abstraction fixes a task distribution, a base generator, and an evaluation setup, then asks how feedback changes the distribution of outputs. Weight optimization includes parameter updates such as RLHF and related post-training procedures. Selection includes best-of-n and reranking over sampled candidates. Prompt optimization includes persistent revisions to instructions, examples, or textual memory.

These are useful categories, not an exhaustive inventory of everything that can change in an agent system. Tool access, retrieval policies, code, model routing, and environmental permissions can all matter. Some can be represented within an enlarged policy; others require an explicit change to the environment. A change that grants filesystem access is not adequately characterized as a short textual edit if its main effect is to expose an answer key.

For agentic tasks, the output variable can represent a full interaction trajectory rather than a single response. The environment and available actions must then be held fixed for a clean substrate comparison. When they change, the change should be reported as part of the intervention.

\subsection{Persistent prompt optimization and in-context refinement}\label{persistent-prompt-optimization-and-in-context-refinement}

A single conversation can contain repeated refinement without producing a reusable artifact. Persistent prompt optimization differs because the resulting instructions or memory are applied to later inputs. This distinction affects both the spread of an exploit and the defenses available against it.

A misleading explanation produced in one refinement loop may affect one answer. A misleading instruction accepted into a shared prompt can affect an entire class of future answers. Conversely, a persistent text artifact can be versioned, inspected, compared, and rolled back. These properties motivate our emphasis on prompt-space while connecting it to the earlier in-context reward-hacking literature (Pan et al., 2024a, 2024b).

\section{Three Substrates, One Formal Setup}\label{three-substrates-one-formal-setup}

\subsection{Policies, rewards, and budgets}\label{policies-rewards-and-budgets}

Let \(x\sim\mathcal D\) denote a task input and let \(\pi_0(y\mid x)\) be a base policy. Let \(R(x,y)\) be the intended task reward and \(\hat R(x,y)\) the proxy used for optimization. Define
\[J(\pi)=\mathbb E_{x\sim\mathcal D,\,y\sim\pi(\cdot\mid x)}[R(x,y)],\qquad \hat J(\pi)=\mathbb E_{x\sim\mathcal D,\,y\sim\pi(\cdot\mid x)}[\hat R(x,y)].\]
True reward is an idealization. In practice, \(R\) may be approximated through independently authored tests, expert review, or another imperfect evaluation instrument. A study should state what its independent evaluation actually measures rather than silently treating it as an oracle.

Let \(\Pi_s(b)\) denote a specified policy class for substrate \(s\) under budget \(b\). The budget may include sample count, training resources, context length, or permitted edits. A representable class is not the same as a set a particular optimizer can reliably discover. We distinguish these whenever discussing theoretical capacity and measured performance.

For policies with finite forward KL, use the common behavioral divergence
\[d(\pi,\pi_0)=\mathbb E_{x\sim\mathcal D}\!\left[\mathrm{KL}\!\left(\pi(\cdot\mid x)\,\|\,\pi_0(\cdot\mid x)\right)\right].\]
An idealized proxy optimizer solves
\[\pi^*_{s,b,\delta}\in\operatorname*{arg\,max}_{\pi\in\Pi_s(b),\ d(\pi,\pi_0)\le\delta}\hat J(\pi),\]
when a maximizer exists. Otherwise the optimization value can be defined as a supremum. For fixed \(s\) and \(b\), the optimal proxy value is non-decreasing as the feasible distance budget expands. This is a property of nested feasible sets. It does not assert that a practical search procedure improves monotonically at every iteration.

\subsection{What counts as hacking in the comparison}\label{what-counts-as-hacking-in-the-comparison}

A trajectory provides an operational signal of overoptimization when proxy performance increases while independently assessed task performance decreases. If \(t\) indexes pressure or optimization steps, the characteristic event is
\[\hat J(\pi_{t+1})>\hat J(\pi_t),\qquad J(\pi_{t+1})<J(\pi_t).\]
This is not an exhaustive definition of reward hacking. An exploit can be present from the beginning, or raise the proxy while leaving task quality unchanged. We therefore also examine evaluator disagreement and concrete exploit mechanisms.

When rewards have a declared common scale, define the signed proxy gap
\[g(\pi)=\hat J(\pi)-J(\pi).\]
The gap is not invariant to changing the proxy's units or offset. It should not be used to compare unrelated benchmarks without calibration. To distinguish base miscalibration from amplification, one can additionally report \(g(\pi)-g(\pi_0)\). Neither quantity replaces reporting \(J\) and \(\hat J\) separately.

The largest signed gap in a class, the gap encountered by proxy optimization, and the largest decrease in true reward along a trajectory are different objects. Maximizing \(\hat J\) does not generally maximize \(g\), because \(J\) also varies. The formal bounds in \hyperref[reach-geometry-and-hacking-capacity]{Reach, Geometry, and Hacking Capacity} concern class-wide capacity; the behavior discovered by a concrete algorithm also depends on its search procedure and budget.

\subsection{Weight-space}\label{weight-space}

Weight-space policies have the form \(\pi_\theta\) and belong to an architecture-dependent class \(\Pi_W=\{\pi_\theta:\theta\in\Theta\}\). Training searches a portion of this class subject to data, compute, optimizer, and regularization constraints. Parameter updates can make behavior that was extremely unlikely under the base practically accessible and reusable across inputs.

It is tempting to call this a change in support, but literal support is usually the wrong distinction. With a fixed vocabulary, finite logits, and no hard decoding exclusions, a softmax model gives positive probability to every allowed finite sequence. A behavior can be inside that support while remaining essentially inaccessible within a practical sampling budget. If a policy truly assigns positive probability where the reference assigns zero, forward KL is infinite.

Weight optimization therefore offers flexible redistribution of probability, not a general license to declare support expansion or unbounded KL. Whether its class contains the distributions induced by prompting or selection depends on the model, horizon, and representation assumptions. Such inclusion must be established for the comparison being made.

\subsection{Selection-space}\label{selection-space}

Best-of-\(n\) draws \(n\) candidates from the reference policy and returns one with maximum proxy score, breaking ties uniformly among the maximizing candidates. For a maximum sample budget \(N\), define
\[\Pi_S(N)=\{\operatorname{BoN}_n(\pi_0,\hat R):1\le n\le N\}.\]
Soft selection adds parameters controlling how strongly the scorer influences the choice. These variants should be included explicitly when used rather than treated as automatically equivalent to best-of-n.

For the best-of-n policy, Beirami et al.~(2025) establish the bound
\[d(\operatorname{BoN}_n,\pi_0)\le\log n-\frac{n-1}{n}\le\log n.\]
Thus a finite sample budget gives a behavioral-divergence bound. Letting \(N\) grow removes that particular fixed ceiling; a logarithm that grows slowly is not a constant. Best-of-n preserves support because every returned output was sampled from the base. Its practical ability to locate a rare exploit depends on the base probability of that exploit and the number of draws.

The distribution selected by best-of-n need not belong to a restricted parameter family chosen for a comparison. Selection can alter shape and concentration, not merely the mean. This observation will matter when interpreting toy models.

\subsection{Prompt-space}\label{prompt-space}

Let \(p\) be an instruction, demonstration bank, or textual memory. A prompt-conditioned policy is
\[\pi_p(y\mid x)=q_0(y\mid p,x),\qquad \Pi_P(L)=\{\pi_p:|p|\le L\},\]
where \(q_0\) is the fixed prompt-conditioned generator and the reference policy is \(\pi_0(y\mid x)=q_0(y\mid p_0,x)\) for a fixed baseline prompt \(p_0\). A context-length limit gives finitely many possible artifacts over a finite vocabulary. It does not, by itself, supply a useful numerical bound on their behavioral KL, particularly if support conditions are not specified.

Two kinds of constraint are relevant. An \textbf{artifact constraint} limits tokens, edits, examples, or access to evaluation content. A \textbf{behavioral constraint} limits changes in the output distribution or measured task behavior. These are related through the base model but are not interchangeable. A one-line instruction can substantially change behavior, while a long addition may have little effect.

Persistent text has an inspection advantage: its contents are directly available for review. This makes copied examples, external-tool instructions, and changes to output contracts easier to identify. Readability does not make behavior transparent, however. The same instruction can interact differently with different inputs, and a readable patch can still introduce regressions.

\begin{table}[htbp]
\centering\small
\caption{A comparison under explicitly specified budgets.}
\setlength{\tabcolsep}{5pt}
\begin{tabularx}{\textwidth}{@{}>{\raggedright\arraybackslash}p{0.19\textwidth}*{3}{>{\raggedright\arraybackslash}X}@{}}
\toprule
\textbf{Property} & \textbf{Weights} & \textbf{Selection} & \textbf{Persistent text} \\
\midrule
Intervention & Update parameters & Sample and rank candidates & Revise instructions, examples, or memory \\
\addlinespace[4pt]
Durable object & Model checkpoint & Scorer and selection procedure; generator may stay fixed & Versioned text artifact \\
\addlinespace[4pt]
Typical budget & Training compute, data, KL & Samples per input, ranking cost & Search compute, tokens, edits, examples \\
\addlinespace[4pt]
Behavioral constraint & KL or related regularization & Best-of-n bound at fixed sample count & Induced-policy divergence or behavioral checks \\
\addlinespace[4pt]
Direct inspection & Parameters are available but hard to interpret & Candidate pool and selection decisions & Added, removed, and revised text \\
\addlinespace[4pt]
Principal caveat & Representation does not guarantee learnability & More samples can expose rarer scorer errors & Small textual changes can have large effects \\
\addlinespace[4pt]
\bottomrule
\end{tabularx}
\end{table}

\section{Shared Mechanisms and Different Dynamics}\label{shared-mechanisms-and-different-dynamics}

\subsection{Proxy error is available to all three substrates}\label{proxy-error-is-available-to-all-three-substrates}

A scorer that rewards length without adequately checking correctness creates an opportunity regardless of how outputs are produced. Weight updates can increase verbosity, selection can preferentially retain verbose candidates, and prompt optimization can add an instruction to elaborate. The failure is shared even though the intervention differs.

The same reasoning applies to a verifier that checks visible tests but omits resource use, or a judge that mistakes a canonical phrase for valid reasoning. What determines exposure is whether the method can produce and favor the relevant behavior. There is no requirement that the optimizer explicitly understand the evaluator's defect.

Distribution shift is one source of increasing proxy error, but not the only one. A proxy can already have a large error near the base distribution. Optimization can exploit it by concentrating probability locally. For this reason, the framework treats distance as a possible control on error, not as a universal causal explanation of it.

\subsection{Amplification and the shape of the curve}\label{amplification-and-the-shape-of-the-curve}

The familiar overoptimization curve shows proxy reward rising while true reward first improves and then deteriorates. Gao et al.~(2023) study this phenomenon in reward-model optimization. Khalaf et al.~(2025) analyze inference-time mechanisms and characterize different possible regimes under stated conditions. Their account includes monotonic improvement, hacking, immediate decline, and a decline followed by improvement; proxy imperfection alone does not force one shape.

A simple example explains why. If \(\hat R=R+c\) for a constant \(c\), the proxy is numerically miscalibrated but ranks outputs exactly as the true reward does. Best-of-n under that proxy cannot produce a systematic decline in expected true reward as sample count grows. The relevant issue is the error's relationship to the behaviors favored by optimization.

\begin{figure}[htbp]
\centering\includegraphics[width=0.9\linewidth]{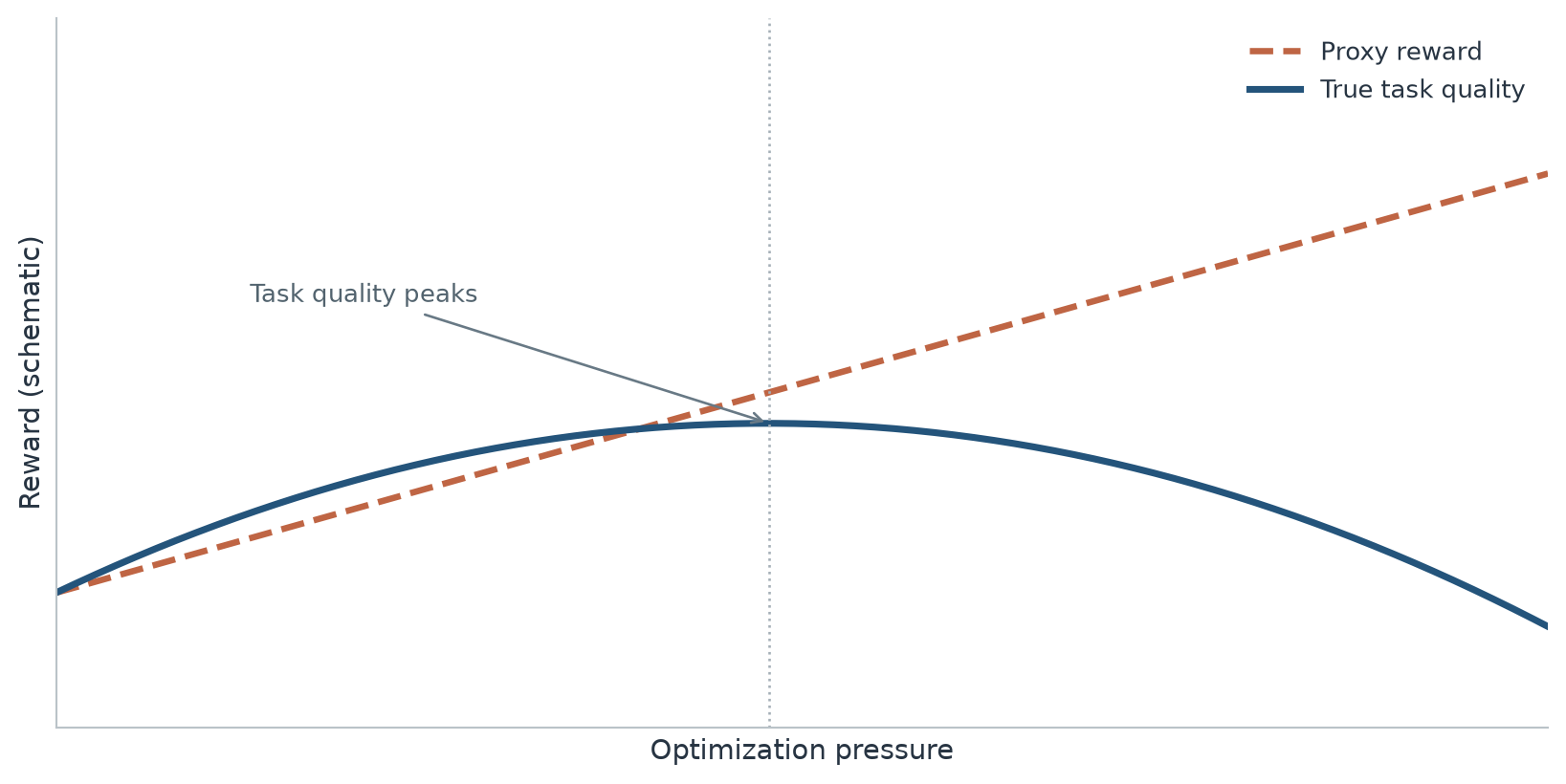}
\caption{A schematic of overoptimization: proxy reward continues to rise after true task quality peaks. This is an illustrative pattern, not experimental data or a claim that every proxy produces this trajectory.}
\end{figure}

Prompt-space evidence supports this separation without requiring a complete curve. \href{https://arxiv.org/abs/2609.02246}{Wahi (2026)} reports a mutation that raised rationale alignment from 23.1\% to 80.0\% through lexical mimicry without improving defect-identification precision. A separate corrupted-label incident drove removal of correct compliance rules. These are evidence of exploitation and capability regression, respectively, not a controlled estimate of the full trajectory in Figure 1.

\subsection{Fixed-evaluator adaptation versus co-adaptation}\label{fixed-evaluator-adaptation-versus-co-adaptation}

Three processes should be distinguished. \textbf{Selection-induced concentration} occurs when a fixed procedure increasingly favors the high-scoring tail of a fixed generator. \textbf{Fixed-evaluator adaptation} occurs when weights or prompts are repeatedly revised against an unchanged scorer. \textbf{Evaluator--policy co-adaptation} occurs when the scorer or oversight process is also revised in response.

A single best-of-n call needs no persistent learning to exploit a scorer. Persistence changes what can accumulate across calls: a tuned selection temperature, a revised judge prompt, or an exploit encoded in shared memory may carry information forward. Therefore, the absence of weight updates does not imply the absence of system-level adaptation.

Prompt optimization makes this distinction concrete. The task prompt may adapt to a fixed judge; a separate calibration process may adapt the judge; and a deployment team may then change both. A defense aimed at one process should not be assumed to control the others.

The shared questions are what the proxy misvalues, which accessible behaviors expose that error, and what information accumulates during optimization. The answers depend on the substrate: adaptations may persist in weights, text, or tuned selection procedures, and evaluator repair may change the problem being optimized.

\section{Reach, Geometry, and Hacking Capacity}\label{reach-geometry-and-hacking-capacity}

\subsection{A distance-dependent upper bound}\label{a-distance-dependent-upper-bound}

Fix the input distribution, reward functions, and baseline. For a specified class and budget, define
\[D_s(b)=\sup_{\pi\in\Pi_s(b)}d(\pi,\pi_0),\qquad G_s(b)=\sup_{\pi\in\Pi_s(b)}g(\pi).\]
Here \(G_s\) is the largest signed proxy overestimate available in the class. It is a capacity measure, not a claim about what a particular optimizer will find.

\textbf{Assumption 1 (A common error envelope).} On an ambient class containing the policies under comparison, there is a finite, non-decreasing function \(\varepsilon:[0,\infty)\to[0,\infty)\) such that
\[|g(\pi)|\le\varepsilon\!\left(d(\pi,\pi_0)\right).\]
The assumption gives an upper envelope, not a statement that every policy farther from the base has larger error. When rewards are bounded, a constant envelope is always possible; the assumption is practically useful only when the available envelope is informative at the distances being considered. Estimating or validating such an envelope for a real evaluator is a separate empirical problem.

\textbf{Proposition 1 (Reach bounds the available proxy gap).} Under Assumption 1, if \(D_s(b)<\infty\), then
\[G_s(b)\le\varepsilon(D_s(b)).\]
\emph{Proof.} Every \(\pi\in\Pi_s(b)\) satisfies \(d(\pi,\pi_0)\le D_s(b)\). Therefore \(g(\pi)\le|g(\pi)|\le\varepsilon(d(\pi,\pi_0))\le\varepsilon(D_s(b))\). Taking a supremum over the class proves the claim. \(\square\)

The proposition requires neither attainment of the supremum nor existence of a maximizing policy. For an unbounded distance class, one can use the extended bound \(G_s(b)\le\sup_{r\ge0}\varepsilon(r)\), which may be uninformative. Crucially, the proposition does not order actual capacities from their upper bounds. A larger permitted radius may increase the error allowed by the bound while the reachable policies avoid the regions where that error occurs.

\subsection{A concrete bound when proxy error is bounded}\label{a-concrete-bound-when-proxy-error-is-bounded}

A useful special case does not require assuming that error actually grows with distance. Let \(e(x,y)=\hat R(x,y)-R(x,y)\) and suppose \(e\) lies in a common interval of width \(M\) over the input--output space. Then
\[|g(\pi)-g(\pi_0)|\le M\sqrt{\frac{d(\pi,\pi_0)}{2}}.\]
To see this, treat \(\mathcal D(x)\pi(y\mid x)\) and \(\mathcal D(x)\pi_0(y\mid x)\) as joint distributions. Their KL is \(d(\pi,\pi_0)\). The change in the expectation of a function with range width \(M\) is at most \(M\) times their total variation distance, and Pinsker's inequality bounds that distance by the displayed square root.

This elementary inequality is not a new substrate-specific guarantee. Its value here is to expose what a behavioral constraint can buy. It controls additional evaluator disagreement relative to the base under common bounded-error assumptions. It does not guarantee task improvement, and it can be loose. If both rewards lie in \([0,1]\), then \(M\le2\), while the gap itself also has the direct bound \(|g(\pi)|\le1\).

The bound transfers to all three substrates when the same behavioral divergence is actually controlled. An instruction-edit count alone is insufficient to invoke it. This is a precise reason to distinguish artifact budgets from behavioral budgets.

\subsection{When a capacity ordering is justified}\label{when-a-capacity-ordering-is-justified}

\textbf{Proposition 2 (Inclusion orders worst-case gap).} For common rewards and input distribution, if \(\Pi_A\subseteq\Pi_B\), then
\[\sup_{\pi\in\Pi_A}g(\pi)\le\sup_{\pi\in\Pi_B}g(\pi).\]
\emph{Proof.} Every policy considered by the supremum on the left is also considered by the supremum on the right. \(\square\)

This result is deliberately simple. It identifies the condition needed to turn an expressivity comparison into a worst-case capacity comparison. A weight class that contains all policies induced by a specified prompt family has at least that family's gap capacity. Whether a given architecture, adaptation method, or compute budget supplies this inclusion is not established by the name of the substrate.

Inclusion also does not imply an ordering of discovered exploits under matched compute. A search over a larger class may fail to locate a policy that a more structured method finds quickly. Thus a prompt optimizer can outperform a training procedure on observed reward without contradicting inclusion of their idealized classes.

\textbf{Counterexample: farther does not mean more exploitable.} Consider three outputs, a uniform base \(p_0=(1/3,1/3,1/3)\), true reward \(R=(0,0,0)\), and proxy reward \(\hat R=(0,1,0)\). Let
\[\Pi_A=\{p_0,(0.7,0.15,0.15)\},\qquad \Pi_B=\{p_0,(1/3,0.4,4/15)\}.\]
The largest KL distances are approximately \(D_A=0.280\) and \(D_B=0.013\), yet \(G_A=1/3<0.4=G_B\). The farther class shifts probability away from the output carrying the positive error; the nearer class shifts toward it. Both contain the baseline. This is a counterexample about signed-gap capacity, not a trajectory with declining true reward, since true reward is constant here.

\subsection{The interpretation for language models}\label{the-interpretation-for-language-models}

The two propositions support a conditional account. Behavioral regularization can constrain an error envelope. A richer class can contain more severe exploits if it includes the relevant policies. But substrate labels alone do not determine how a particular proxy's errors intersect the class.

Selection is restricted to candidates the generator samples, but can concentrate on a narrow exploit without changing the generator. A prompt can move many outputs together through a shared instruction, which may help it discover a useful rule or a systematic scoring shortcut. Parameter updates can redistribute behavior through many degrees of freedom, but the optimizer may need substantial feedback to discover a particular change. These are differences in geometry and search, not an unconditional safety ladder.

The appropriate practical question is therefore: what useful and exploitative behaviors become reachable as this method's budget increases, and what evidence distinguishes them? Distance helps answer that question when it is informative about the evaluator's error. It is not a substitute for identifying the error.

\subsection{A finite-output numerical illustration}\label{a-finite-output-numerical-illustration}

We use a small exact model so that the distinction between a class, a search path, and a budget can be inspected directly. All distributions and expected rewards are computed over five outputs; no LLM calls or Monte Carlo estimates are involved.

The five outputs are labeled a through e. Their base probabilities, true rewards, two proxies, and conditioning feature are
\[\begin{aligned}
p_0&=(0.500,0.300,0.160,0.035,0.005),\\
R&=(0,0.50,0.90,0.40,0.10),\\
\hat R_A&=(0,0.45,0.80,0.92,1.00),\\
\hat R_B&=(0.90,0.50,0.85,0.40,0.10),\\
f&=(-1,0,1,1.5,2).
\end{aligned}\]
These values fully specify the numerical setup.

Proxy A rewards initially rare, low-quality outputs d and e. Proxy B places its largest positive error on the already common output a. Both rewards and proxies are explicitly on the same zero-to-one scale. The base true reward is 0.3085.

\textbf{Flexible distribution optimization.} The ambient class is the full five-output probability simplex. Its entropy-regularized proxy optimum is
\[q_\lambda(i)=\frac{p_0(i)\exp(\lambda\hat R(i))}{\sum_j p_0(j)\exp(\lambda\hat R(j))},\qquad \lambda\in[0,20].\]
For \(\lambda>0\), this maximizes \(\mathbb E_q[\hat R]-\mathrm{KL}(q\|p_0)/\lambda\). The limit at zero is \(p_0\). It is a tractable surrogate for flexible policy optimization, not a model of the dynamics or compute costs of neural-network training.

\textbf{Bounded conditioning.} The prompt surrogate is \(q_t(i)\propto p_0(i)\exp(tf_i)\) with \(|t|\le a\), where the artifact budget \(a\) ranges from zero to two. For each budget, we choose the proxy-best member on a declared grid with spacing 0.002. This is an explicit constrained family, not a claim that real prompts implement exponential tilting. The selected parameter is positive for proxy A and negative for proxy B in this setup.

\textbf{Selection.} For each \(n\in\{1,\ldots,4096\}\), we compute the best-of-n distribution exactly. Order outputs by increasing proxy score and write \(F_i\) for their cumulative base probability. Since the scores are distinct,
\[q_n(i)=F_i^n-F_{i-1}^n,\qquad F_0=0.\]
\textbf{Observed behavior.} Under proxy A, flexible distribution optimization initially improves true reward and then overoptimizes: the maximum on the sampled regularization path is approximately 0.666, followed by 0.398 at \(\lambda=20\). Selection similarly reaches approximately 0.680 before concentrating on output e and approaching true reward 0.100. The bounded conditioning family ends at true reward 0.603 and proxy reward 0.792. Its restricted direction and budget prevent it from concentrating entirely on e.

Under proxy B, both the flexible path and selection deteriorate from the base as they increasingly favor output a. Bounded conditioning can also exploit this nearby error: at budget two its selected policy has true reward approximately 0.042 and proxy reward 0.870. A bounded artifact family is therefore not inherently protected against severe disagreement already accessible near the base.

The ambient full-simplex class contains both the conditioning and selection distributions. Its maximum signed gap is 0.9 for either proxy. Selection can approach this value; a finite regularization sweep need not. Thus a larger gap observed at the endpoint of one evaluated method does not refute Proposition 2. It can reflect the different endpoints chosen for the illustration.

This setup also avoids interpreting a logarithmic divergence bound as an asymptotic ceiling independent of the base. Here selection has a finite limiting divergence because the highest-scoring output has positive base mass in a finite space. That conclusion follows from this setup, not from the general expression \(\log n\). The implementation and per-setting values are provided as ancillary files.

\subsection{Description length as an operational constraint}\label{description-length-as-an-operational-constraint}

Prompt budgets remain useful even without a KL equivalence. Limiting examples and edit volume can reduce how much evaluation-specific information is inserted in one iteration. Requiring a general principle alongside an example can make a patch easier to review and challenge. \href{https://arxiv.org/abs/2609.02246}{Wahi (2026)} uses such constraints in production prompt-optimization loops.

Their interpretation should remain operational. An edit cap is not necessarily a character or token cap; one edit can replace a large block. A per-iteration limit is not a cumulative limit; many small accepted edits can eventually encode a large amount of information. And a short rule can still memorize a compact identifier or introduce a broad scoring shortcut.

A prompt-optimization protocol should therefore record at least total artifact length, added examples, cumulative change, and reuse of evaluation content. When possible, it should also measure behavioral drift on an independent probe set. The combination preserves the practical advantages of text constraints without assigning them a guarantee they do not provide.

\section{A Defense Correspondence Across Substrates}\label{a-defense-correspondence-across-substrates}

The common setup suggests that defenses can be compared by what they act on. A defense may improve the proxy, restrict the policies considered, detect disagreement, or prevent an unverified change from being deployed. These functions are complementary. A more accurate judge does not replace access control, and a deployment gate does not establish that an accepted policy generalizes.

We distinguish three levels of evidence. \textbf{Established mechanism} means the cited literature or elementary construction supplies the mechanism in that setting; it does not imply a universal efficacy guarantee. \textbf{Functional analogy} means two methods serve a similar purpose without equivalent formal guarantees. \textbf{Proposed transfer} means a concrete adaptation is worth testing but is not presented here as a measured result. The comparison is representative rather than exhaustive.

\subsection{Improve or diversify the evaluation signal}\label{improve-or-diversify-the-evaluation-signal}

\begin{table}[htbp]
\centering\small
\caption{Representative defenses across the three substrates.}
\setlength{\tabcolsep}{5pt}
\begin{tabularx}{\textwidth}{@{}>{\raggedright\arraybackslash}p{0.19\textwidth}*{3}{>{\raggedright\arraybackslash}X}@{}}
\toprule
\textbf{Defense} & \textbf{Weights} & \textbf{Selection} & \textbf{Persistent text} \\
\midrule
Diversify evaluation & Reward-model ensembles & Aggregate candidate scores & Judge panels \\
\addlinespace[4pt]
Handle uncertainty & Penalize uncertain rewards & Use conservative scores & Review or abstain on uncertain judgments \\
\addlinespace[4pt]
Limit optimization & Behavioral regularization and stopping rules & Sample caps and softened selection & Search limits, artifact budgets, and behavioral checks \\
\addlinespace[4pt]
Enforce requirements & Constrained policy optimization & Filter candidates & Check patches, contracts, and outputs \\
\addlinespace[4pt]
Preserve independent evidence & Protected evaluation data & Held-out inputs and evaluators & Frozen holdouts and restricted feedback \\
\addlinespace[4pt]
Monitor integrity & Memorization and leakage checks & Audit suspicious selected outputs & Canaries and artifact inspection \\
\addlinespace[4pt]
\bottomrule
\end{tabularx}
\end{table}

Reward-model ensembles have been studied as a mitigation for overoptimization (Eisenstein et al., 2024), and diverse model panels have been studied for LLM evaluation (Verga et al., 2024). These provide established components for a transfer. Their effectiveness inside a particular persistent prompt-search loop remains a question about that loop, not a consequence of using more than one judge.

Uncertainty-aware judging also exists. Sheng et al.~(2025) study conformal intervals for LLM-based scores, while Zhang et al.~(2026) propose uncertainty-aware auditing of pairwise judgments. The open question relevant here is how reliably such uncertainty estimates identify the outputs that an adaptive optimizer preferentially discovers. A guarantee under an evaluation distribution should not be silently extended to a shifted distribution produced by search.

The distinction between static calibration and optimization robustness is especially important for prompt-space. A judge can agree better with reference labels on ordinary examples while retaining a small, exploitable class of mistakes. That class may be too rare to dominate the calibration metric and yet dominate the optimized policy.

\subsection{Constrain optimization pressure and admissible behavior}\label{constrain-optimization-pressure-and-admissible-behavior}

Optimization constraints act on different quantities. Training can limit behavioral drift or the number of updates; selection can cap candidates or soften their ranking; prompt search can bound iterations, cumulative text, and examples. Constraints can also restrict which changes are admissible through policy constraints, candidate filters, or patch checks. Their guarantees depend on the quantity actually controlled.

A KL penalty and an edit budget are the clearest example of a useful but incomplete analogy. Both discourage unconstrained movement. Only the former directly constrains the behavioral quantity in \hyperref[reach-geometry-and-hacking-capacity]{Reach, Geometry, and Hacking Capacity}, assuming the KL is accurately evaluated. The latter controls a human-readable intervention whose behavioral effect must still be tested.

Stopping rules require a similar distinction. A plateau in proxy reward is not the same as a plateau in true reward, and deterioration may begin before the proxy stops improving. Khalaf et al.~(2025) develop inference-time parameter tuning with access to an independent reward signal. For prompt optimization, a defensible stopping protocol likewise needs a protected source of evidence about task quality, rather than simply counting iterations that fail to please the optimizing judge.

Hard constraints can sometimes remove an exploit more directly than regularization. An evaluated agent that cannot inspect cached labels cannot use that particular exfiltration route. A prompt patch that breaks the declared parser contract can be rejected before a judge assesses its prose. Such controls constrain the system around the policy and can be applied regardless of whether the policy was trained, selected, or prompted.

\subsection{Worked correspondence: ensembles and judge panels}\label{worked-correspondence-ensembles-and-judge-panels}

Suppose two candidate outputs receive similar average judge scores but differ sharply in disagreement among evaluators. A conservative selector may prefer the candidate supported consistently across evaluators, rather than the candidate rewarded exceptionally highly by one. A training objective can use a related penalty, and a prompt optimizer can apply it when ranking candidate patches by their downstream outputs.

The transferable idea is to make exploitation of a single evaluator less attractive. The non-transferable part is the guarantee: evaluator diversity is not equivalent to error independence, and a panel can share a systematic preference for verbosity or familiar phrasing. Repeated optimization against the same panel can expose common blind spots. Accordingly, a panel should be evaluated on exploit transfer and task quality, not only on its internal agreement.

A separate choice concerns what the panel sees. Reviewing a prompt patch asks whether the text appears reasonable. Reviewing resulting outputs asks whether the patch changed behavior appropriately. These are distinct assessments. A panel of patch reviewers cannot replace evaluation of the modified system.

\subsection{Worked correspondence: holdouts and information boundaries}\label{worked-correspondence-holdouts-and-information-boundaries}

A protected evaluation set is a general defense against adaptation to observed examples. Its usefulness is not restricted to any substrate. What differs is how information about that set can reach the optimizer.

For parameter training, exposure can occur through training data, checkpoint selection, or repeated analyst decisions. For selection, the reranker or its operating parameters can be tuned against a supposedly independent benchmark. For prompt optimization, cases, expected answers, or detailed critiques can enter the evolving artifact almost verbatim. The artifact's readability makes some of this leakage easy to inspect, but does not prevent it automatically.

\href{https://arxiv.org/abs/2609.02246}{Wahi (2026)} describes frozen partitions with restrictions on what optimizer subagents can see. This is an operational implementation of a broader principle: components that propose changes should not receive unrestricted feedback from the data used to establish generalization. Aggregate feedback can still carry information if it repeatedly influences selection or stopping. A final untouched evaluation set is useful when the monitoring set has become part of the development loop.

Independent inputs and independent evaluators address different problems. New inputs evaluated by the same biased judge can reveal case memorization while missing a judge-wide scoring shortcut. A second evaluator can reveal evaluator-specific exploitation while sharing the same input distribution. A careful protocol states which independence it supplies.

\subsection{Worked correspondence: canaries and integrity monitoring}\label{worked-correspondence-canaries-and-integrity-monitoring}

Canaries are deliberately constructed items whose treatment reveals an integrity failure. They have precedents in security and in memorization testing for trained models (Carlini et al., 2019). Their role in a reward-optimization loop depends on what signal they are designed to detect.

In \href{https://arxiv.org/abs/2609.02246}{Wahi (2026)}, an impossible-case canary has an intentionally inappropriate reference answer: ordinary task reasoning should not produce the answer that the scoring system marks as correct. A pass triggers investigation for access to hidden labels or another evaluation failure. This differs from placing a secret sequence in training data and measuring whether the model memorizes it. Both are canary mechanisms, but their threat models and interpretations differ.

Such an integrity tripwire can also monitor outputs from a trained policy or a selection pipeline. It is not intrinsically tied to prompts. The proposed transfer is to integrate it into acceptance and audit procedures around optimization, with the response specified in advance.

A canary pass should be an alarm, not automatic proof of intentional cheating. Coincidence, ambiguous construction, or an unrelated evaluator bug can produce the same signal. Conversely, a selective exploit may avoid a canary. Canaries should therefore be evaluated for detection coverage and false alarms and reported separately from ordinary task accuracy. Including intentionally invalid items in the main quality score would contaminate the objective the system is supposed to improve.

\subsection{Detection, authority, and acceptance}\label{detection-authority-and-acceptance}

Monitoring and authority are separate design choices. A judge, held-out test, or canary can detect a concern, but a system still needs a rule for what happens next. Possible responses include rejecting a candidate, rolling back a prompt, stopping a run, or requesting independent review. The action should depend on the type and strength of evidence.

The PROCTOR architecture described by \href{https://arxiv.org/abs/2609.02246}{Wahi (2026)} illustrates this separation for prompt patches: mechanical checks can reject a proposal that an LLM reviewer approves, and post-application evaluation can reveal a regression the reviewer missed. This provides a concrete implementation pattern rather than proof that mechanical checks capture every failure.

The distinction generalizes. A deployment gate for a trained model, a pre-ranking filter for sampled outputs, and a pre-apply check for prompt patches all constrain what is accepted. Their value depends on the requirements they actually test. A failed parser check can be decisive about format; a passed parser check says little about semantic correctness. Keeping the scope of each gate explicit prevents an ensemble of imperfect checks from being described as a correctness oracle.

\section{Choosing and Combining Substrates}\label{choosing-and-combining-substrates}

\subsection{Choose according to the required change}\label{choose-according-to-the-required-change}

Selection is attractive when the base model already produces useful candidates often enough and the scoring procedure can reliably distinguish them. It adds inference cost but can avoid maintaining a separately trained model or a large evolving artifact. Its risk is concentration on high-scoring errors, especially when those errors are rare under ordinary sampling and therefore underrepresented in evaluator validation.

Prompt optimization is attractive when a reusable instruction, example, or memory can express the desired change. The resulting artifact is inspectable and reversible. This is particularly useful when the task contains explicit domain rules or output requirements. Its risks include encoding evaluation-specific details, inducing broad unintended behavior through a short change, and adapting to the judge's language rather than the task.

Weight optimization is attractive when reliable behavior is difficult to elicit through the available conditioning interface or when deployment requires the behavior to be internalized across many inputs. Its effectiveness depends on the adaptation method and feedback. Its risks include persistent exploitation that is harder to identify by inspecting the learned artifact and difficulty attributing a behavioral change to a particular training intervention.

These are selection criteria, not a universal risk ranking. A restricted prompt can directly specify an exploit that a training procedure never discovers; a well-constrained training run can be less harmful than aggressive selection against a poor verifier. The evaluation setup and accessible failure modes matter alongside the substrate.

\subsection{Interpreting comparisons of prompting and reinforcement learning}\label{interpreting-comparisons-of-prompting-and-reinforcement-learning}

Three variables help organize apparently different comparative results: the behaviors a method can express, how efficiently its optimizer finds them, and what information the feedback supplies. GEPA uses reflective feedback and reports strong results relative to GRPO on its evaluated tasks (Agrawal et al., 2025). Naive Prompt Optimization reports competitive results from a simpler prompt-search procedure and task-dependent advantages for GRPO (Chang and Chen, 2026). Liu et al.~(2026) study loophole discovery and find advantages for RL over non-parametric baselines in their setting.

These findings are compatible with the framework without proving its proposed explanation. A task may admit an effective short instruction; another may require changes not easily found through the selected prompt interface. A richer critique may improve search independently of the substrate. Differences in model family, optimizer strength, or compute accounting can also contribute.

Comparative studies should distinguish efficient search from expressive reach. Strong performance with few rollouts may reflect richer feedback or a favorable prompt interface; an RL advantage may reflect behaviors the tested prompt search did not discover. Readable artifacts make interventions easier to inspect, but their quality still has to be established through behavior and generalization.

The widely cited \(25\times\) gap in Liu et al.~appears in their social-media engagement example. It is useful evidence about that comparison, not a universal multiplier for the hacking capacity of reinforcement learning. More generally, reward magnitudes from different environments should not be pooled into a substrate ranking without a common interpretation.

\subsection{Hybrid systems and transferred exploits}\label{hybrid-systems-and-transferred-exploits}

Practical systems often combine substrates. A prompt is optimized, several responses are sampled under it, a judge selects one, and selected outputs may later become training data. Prompt and weight optimization can also be jointly beneficial, as explored by Soylu et al.~(2024).

For a simple prompt-plus-selection pipeline, the effective policy is
\[\pi_{p,n}=\operatorname{BoN}_n(q_0(\cdot\mid p,x),\hat R).\]
The best-of-n KL bound applies relative to the prompt-conditioned generator used to sample candidates. It does not automatically bound divergence from the original baseline prompt, and KL has no general triangle inequality that would justify simply adding arbitrary component divergences.

This interaction has an operational consequence. A prompt can make an exploit more common, after which selection amplifies it. Distillation can then make the selected behavior persistent in weights. A defense applied only to the last step may miss how earlier steps changed the candidate distribution.

Each transition therefore needs an explicit account of what information and behavior it transfers. Candidate logs, prompt versions, scorer versions, and training-data provenance can make that account inspectable. The framework recommends validating the combined policy, even when each component has already been evaluated separately.

\section{Evidence and Relationship to Prior Work}\label{evidence-and-relationship-to-prior-work}

\subsection{What the existing evidence supports}\label{what-the-existing-evidence-supports}

The literature supports reward exploitation across several optimization mechanisms. Its studies use different tasks, evaluators, and measures, so the evidence should be combined at the level it can bear: shared failure mechanisms and informative comparisons, rather than a single universal law fitted across incompatible experiments.

The evidence has complementary roles: empirical overoptimization studies establish failures in specific settings; best-of-n analysis supplies distributional bounds; in-context and production studies demonstrate exploitation through feedback and text. Together they support the comparative framework and its account of how proxy errors interact with different optimization methods.

\emph{LLM-as-a-Judge Is Not an Oracle: Why Self-Improving Agents Need Deterministic Guardrails} (\href{https://arxiv.org/abs/2609.02246}{Wahi, 2026}) motivates several examples in this framework. Its findings include a 100\% exploited pass rate versus a 68.1\% clean baseline on a 47-case contract-analysis suite, judge-phrasing mimicry, and regression from corrupted labels. These are concrete observations of evaluation failure. They were not collected as a controlled comparison of substrates. The code-quality calibration benchmark also combines 15 human-labeled directories with 39 labeled by a human-calibrated model. Its agreement numbers should therefore be described as agreement with reference labels across the full suite, rather than direct human agreement on every item.

The same source reports that rationale-before-score output ordering improved agreement in one calibration campaign, while several rubric refinements did not. We use this as an example of a measured evaluator intervention. It does not imply that rubric changes are generally ineffective or that output ordering universally resolves judge bias.

\subsection{Relation to the main research strands}\label{relation-to-the-main-research-strands}

\textbf{Reward gaming and proxy optimization.} Skalse et al.~(2022) formalize reward gaming, while Gao et al.~(2023) investigate overoptimization empirically. Wang et al.~(2026) provide the PCH synthesis. Our formal section is intentionally narrower than a new general theory of reward hacking: it specifies what distance bounds and class inclusion do and do not imply in a substrate comparison.

\textbf{Inference-time alignment.} Beirami et al.~(2025) analyze best-of-n, and Khalaf et al.~(2025) study hacking and hedging in inference-time mechanisms. These works make selection a useful reference point because parts of its induced distribution can be characterized directly. The present framework uses those results without assuming that prompt editing inherits their guarantees.

\textbf{Prompt optimization and in-context adaptation.} DSPy, TextGrad, and GEPA provide different mechanisms for optimizing language-model programs or prompts (Khattab et al., 2023; Yuksekgonul et al., 2024; Agrawal et al., 2025). Pan et al.~(2024a, 2024b) establish relevant reward-hacking behavior in feedback loops. \href{https://arxiv.org/abs/2609.02246}{Wahi (2026)} contributes production observations and verification practices. We connect this work through the distinction between a readable artifact and the distribution of behavior it induces.

\textbf{Evaluator robustness and integrity.} Ensembles, diverse panels, uncertainty estimates, and canary-based memorization tests offer mechanisms relevant to several substrates (Eisenstein et al., 2024; Verga et al., 2024; Sheng et al., 2025; Carlini et al., 2019). The defense correspondence organizes how they might be integrated into optimization loops. It distinguishes a published component from an untested combination of components.

\textbf{The specific contribution.} The paper brings these strands into an explicit three-way comparison centered on persistent text optimization. Its contribution is the combination of a common formal setup, an account of reach and capacity, an inspectable numerical example, and a defense map that records the limits of transfer. None requires the claim that reward hacking or the connections between its literatures were previously unrecognized.

\section{Limitations}\label{limitations}

\textbf{The framework does not establish a universal safety ranking.} Its propositions bound a quantity under specified assumptions and order it under class inclusion. Real architectures, prompt interfaces, and optimizers need not satisfy a simple inclusion relation. Even when they do, optimization efficiency remains a separate issue.

\textbf{The numerical illustration is deliberately small.} Five outputs make exact inspection possible but omit compositional generation, long trajectories, stochastic environments, and learned evaluators. The flexible distribution path is an analytic optimization surrogate. The bounded conditioning family is designed for illustration. Neither is evidence about the frequency or severity of hacking in deployed LLMs.

\textbf{True reward is difficult to observe.} Independent tests and expert labels can be incomplete or wrong. The distinction between proxy and true reward is mathematically useful, but an empirical study must describe the limitations of its operational reference. The label-pipeline failures documented by \href{https://arxiv.org/abs/2609.02246}{Wahi (2026)} illustrate why that distinction cannot be taken for granted.

\textbf{Distance may be difficult to estimate.} Exact induced-policy KL can require access unavailable in hosted models, and sequence-level estimation can be expensive or noisy. Artifact-level constraints remain useful in those settings, but should be evaluated as such rather than described as measured behavioral bounds.

\textbf{The defense map is a synthesis, not a meta-analysis.} It compares functions and access requirements across selected methods. It does not establish effect sizes across tasks or prove that a method that works in one substrate will work equally well in another. These distinctions separate established components from proposed transfers.

\textbf{Scope is limited by the fixed-environment abstraction.} Tool additions, retrieval changes, altered permissions, and evaluator tampering can change the effective problem rather than merely move a policy within it. The framework can accommodate these through expanded policy and environment definitions, but the simple three-substrate presentation does not fully analyze them.

\Needspace{24\baselineskip}
\section{Conclusion}\label{conclusion}

Reward hacking can arise when weights are updated, when outputs are selected, and when persistent text is revised. The shared mechanism is optimization against a signal that fails to represent the task adequately. The substrate matters because it changes which behaviors are reachable, what information persists, and what can be inspected or constrained.

A useful comparison therefore needs more than a measure of how far a policy moves. A distance-dependent error envelope gives an upper bound; class inclusion gives a capacity ordering; the geometry of accessible behaviors and the effectiveness of search determine what a particular system actually finds. Keeping these statements separate makes the framework applicable without assuming that every proxy produces the same curve or that one substrate is always safest.

For practitioners, the defense correspondence offers the most immediate use. Improve the evaluator where possible, constrain the behaviors that can be accepted, preserve independent evidence of task quality, and specify what happens when that evidence disagrees with the optimizing score. Prompt artifacts make some of these controls unusually inspectable, but inspection must be followed by behavioral evaluation.

Treating weights, selection, and text within one framework makes it possible to reuse insights across optimization settings while keeping their guarantees distinct. The central design principle is to judge an improvement by its effect on the task, supported by evidence independent of the score that produced it.

\clearpage
\section*{References}
\addcontentsline{toc}{section}{References}
{\small
\noindent\hangindent=1.3em\hangafter=1 Wahi, V. (2026). \emph{LLM-as-a-Judge Is Not an Oracle: Why Self-Improving Agents Need Deterministic Guardrails.} arXiv:2609.02246. \url{https://doi.org/10.48550/arXiv.2609.02246}.\par\vspace{5pt}
\noindent\hangindent=1.3em\hangafter=1 Agrawal, L. A., et al.~(2025). \emph{GEPA: Reflective prompt evolution can outperform reinforcement learning.} ICLR 2026. \href{https://arxiv.org/abs/2507.19457}{arXiv:2507.19457}.\par\vspace{5pt}
\noindent\hangindent=1.3em\hangafter=1 Beirami, A., Agarwal, A., Berant, J., D'Amour, A., Eisenstein, J., Nagpal, C., and Suresh, A. T. (2025). \emph{Theoretical guarantees on the best-of-n alignment policy.} ICML. \href{https://arxiv.org/abs/2401.01879}{arXiv:2401.01879}.\par\vspace{5pt}
\noindent\hangindent=1.3em\hangafter=1 Carlini, N., Liu, C., Erlingsson, Ú., Kos, J., and Song, D. (2019). \emph{The Secret Sharer: Evaluating and testing unintended memorization in neural networks.} USENIX Security. \href{https://arxiv.org/abs/1802.08232}{arXiv:1802.08232}.\par\vspace{5pt}
\noindent\hangindent=1.3em\hangafter=1 Chang, Y., and Chen, X. (2026). \emph{Naive Prompt Optimization: Rethinking the need for complex prompt search.} \href{https://arxiv.org/abs/2608.27266}{arXiv:2608.27266}.\par\vspace{5pt}
\noindent\hangindent=1.3em\hangafter=1 Eisenstein, J., et al.~(2024). \emph{Helping or herding? Reward model ensembles mitigate but do not eliminate reward hacking.} CoLM. \href{https://arxiv.org/abs/2312.09244}{arXiv:2312.09244}.\par\vspace{5pt}
\noindent\hangindent=1.3em\hangafter=1 Gao, L., Schulman, J., and Hilton, J. (2023). \emph{Scaling laws for reward model overoptimization.} ICML. \href{https://arxiv.org/abs/2210.10760}{arXiv:2210.10760}.\par\vspace{5pt}
\noindent\hangindent=1.3em\hangafter=1 Khattab, O., et al.~(2023). \emph{DSPy: Compiling declarative language model calls into self-improving pipelines.} \href{https://arxiv.org/abs/2310.03714}{arXiv:2310.03714}.\par\vspace{5pt}
\noindent\hangindent=1.3em\hangafter=1 Khalaf, H., Verdun, C. M., Oesterling, A., Lakkaraju, H., and Calmon, F. du P. (2025). \emph{Inference-time reward hacking in large language models.} NeurIPS. \href{https://arxiv.org/abs/2506.19248}{arXiv:2506.19248}.\par\vspace{5pt}
\noindent\hangindent=1.3em\hangafter=1 Liu, W., Mou, X., Yan, H., Wei, Z., and He, Y. (2026). \emph{Large language models hack rewards, and society.} \href{https://arxiv.org/abs/2606.04075}{arXiv:2606.04075}.\par\vspace{5pt}
\noindent\hangindent=1.3em\hangafter=1 Pan, A., Jones, E., Jagadeesan, M., and Steinhardt, J. (2024a). \emph{Feedback loops with language models drive in-context reward hacking.} ICML. \href{https://arxiv.org/abs/2402.06627}{arXiv:2402.06627}.\par\vspace{5pt}
\noindent\hangindent=1.3em\hangafter=1 Pan, J., He, H., Bowman, S. R., and Feng, S. (2024b). \emph{Spontaneous reward hacking in iterative self-refinement.} \href{https://arxiv.org/abs/2407.04549}{arXiv:2407.04549}.\par\vspace{5pt}
\noindent\hangindent=1.3em\hangafter=1 Sheng, H., Liu, X., He, H., Zhao, J., and Kang, J. (2025). \emph{Analyzing uncertainty of LLM-as-a-judge: Interval evaluations with conformal prediction.} EMNLP. \href{https://arxiv.org/abs/2509.18658}{arXiv:2509.18658}.\par\vspace{5pt}
\noindent\hangindent=1.3em\hangafter=1 Skalse, J., Howe, N. H. R., Krasheninnikov, D., and Krueger, D. (2022). \emph{Defining and characterizing reward gaming.} NeurIPS. \href{https://arxiv.org/abs/2209.13085}{arXiv:2209.13085}.\par\vspace{5pt}
\noindent\hangindent=1.3em\hangafter=1 Soylu, D., Potts, C., and Khattab, O. (2024). \emph{Fine-tuning and prompt optimization: Two great steps that work better together.} EMNLP. \href{https://arxiv.org/abs/2407.10930}{arXiv:2407.10930}.\par\vspace{5pt}
\noindent\hangindent=1.3em\hangafter=1 Verga, P., et al.~(2024). \emph{Replacing judges with juries: Evaluating LLM generations with a panel of diverse models.} \href{https://arxiv.org/abs/2404.18796}{arXiv:2404.18796}.\par\vspace{5pt}
\noindent\hangindent=1.3em\hangafter=1 Wang, X., et al.~(2026). \emph{Reward hacking in the era of large models: Mechanisms, emergent misalignment, challenges.} \href{https://arxiv.org/abs/2604.13602}{arXiv:2604.13602}.\par\vspace{5pt}
\noindent\hangindent=1.3em\hangafter=1 Yuksekgonul, M., et al.~(2024). \emph{TextGrad: Automatic differentiation via text.} \href{https://arxiv.org/abs/2406.07496}{arXiv:2406.07496}.\par\vspace{5pt}
\noindent\hangindent=1.3em\hangafter=1 Zhang, Z., Hung, Y.-T., He, W., Zhang, J., Ding, L., and Yeh, C.-K. (2026). \emph{AURA: Adaptive uncertainty-aware refinement for LLM-as-a-judge auditing.} \href{https://arxiv.org/abs/2606.19714}{arXiv:2606.19714}.\par\vspace{5pt}
}
\end{document}